\documentclass[final]{cvpr}

\usepackage{times}
\usepackage{epsfig}
\usepackage{graphicx}
\usepackage{amsmath}
\usepackage{amssymb}
\usepackage{float}
\usepackage{multirow}
\usepackage{subfig}
\usepackage{wrapfig}

\usepackage{mathtools}% Added by SH, for the NORM symbol.
\DeclarePairedDelimiter{\norm}{\lVert}{\rVert} % Added by SH, for the NORM symbol.
\DeclareMathOperator{\E}{\mathbb{E}} % Added by SH, for the Expectation symbol.
\usepackage[dvipsnames]{xcolor} % Added by SH, for color typing.

\usepackage[pagebackref=true,breaklinks=true,colorlinks,bookmarks=false]{hyperref}

\def\cvprPaperID{8473} % *** Enter the CVPR Paper ID here
\def\confYear{CVPR 2021}

\begin{document}

%%%%%%%%% TITLE
\title{LaPred: Lane-Aware Prediction of Multi-Modal Future Trajectories of Dynamic Agents}

\author{ByeoungDo Kim\textsuperscript{\rm 1} \qquad 
Seong Hyeon Park\textsuperscript{\rm 1} \qquad
Seokhwan Lee\textsuperscript{\rm 1} \qquad
Elbek Khoshimjonov\textsuperscript{\rm 1} \\
Dongsuk Kum\textsuperscript{\rm 2} \qquad
Junsoo Kim\textsuperscript{\rm 3} \qquad
Jeong Soo Kim\textsuperscript{\rm 3} \qquad
Jun Won Choi\textsuperscript{\rm 1}\thanks{
Corresponding author: Jun Won Choi} \\
{\textsuperscript{\rm 1} Hanyang University} \qquad
{\textsuperscript{\rm 2} Korea Advanced Institute of Science and Technology} \\
{\textsuperscript{\rm 3} Hyundai Motor Company} \\
{\tt\small \{bdkim,shpark,shlee,elbek\}@spa.hanyang.ac.kr} \qquad
{\tt\small dskum@kaist.ac.kr} \\
{\tt\small  \{Junsoo.Kim,jskim1074\}@hyundai.com} \qquad
{\tt\small junwchoi@hanyang.ac.kr}
}

\maketitle

%%%%%%%%% ABSTRACT
\begin{abstract}
   In this paper, we address the problem of predicting the future motion of a dynamic agent (called a target agent) given its current and past states as well as the information on its environment.
   It is paramount to develop a prediction model that can exploit the contextual information in both static and dynamic environments surrounding the target agent and generate diverse trajectory samples that are meaningful in a traffic context.
   We propose a novel prediction model, referred to as the {\it lane-aware prediction} (LaPred) network, which uses the instance-level lane entities extracted from a semantic map to predict the multi-modal future trajectories. 
   For each lane candidate found in the neighborhood of the target agent, LaPred extracts the joint features relating the lane and the trajectories of the neighboring agents. Then,  the features for all lane candidates are fused with the attention weights learned through a self-supervised learning task that identifies the lane candidate likely to be followed by the target agent.
   Using the instance-level lane information, LaPred can produce the trajectories compliant with the surroundings better than 2D raster image-based methods and generate the diverse future trajectories given multiple lane candidates. The experiments conducted on the public {\it nuScenes} dataset and {\it Argoverse} dataset demonstrate that the proposed LaPred method significantly outperforms the existing prediction models, achieving state-of-the-art performance in the benchmarks.
\end{abstract}

%%%%%%%%% BODY TEXT
\section{Introduction} \label{introduction}
Predicting the future motion of a dynamic agent given its past trajectory is crucial for self-driving robots and vehicles to conduct path planning and  collision avoidance. However, predicting motion in realistic environments is challenging because the motion of the dynamic agent is determined by various indirectly observed factors, including the agent's intention, the static environment around the agent, and the interaction with other agents. Such uncertainties in prediction tasks entail multiple plausible trajectories that an agent can take to reach its intended goals. Specifically, the distribution of an agent’s future trajectory will be multi-modal in that the agent can exhibit different maneuvers (e.g., right turn, left turn, or straight) for given specific scenes (e.g., four-way crossroad), or change the lanes and adjust its speed interacting with other agents. Therefore,  a prediction model should suggest more than one plausible trajectory samples that follow the distribution for the given situation.

To predict such a diverse set of trajectories, the model must understand the environmental context consisting of social patterns in temporal motion as well as observations for static scene environment via sensors or semantic maps. Therefore, the model’s ability to extract and meaningfully represent such multiple cues is crucial. Deep neural networks (DNN) suit this task well, owing to their large capacity and capability of end-to-end learning representation. Previous art has presented sophisticated architectures for interaction modeling \cite{slstm, trivedi_ConvSP, soft+attn, sgan, trajectron, bigat,desire,shpark_diverse,datadriven_interaction,sophie,trajectronpp,mfp,SS-LSTM, matf}, static-scene processing \cite{multipath, mtp, wimp, desire,lanegcn,covernet, r2p2, sophie, carnet, mapadaptive,matf}, and multi-modal trajectory modeling \cite{multipath,mtp,sgan,diversity,desire,shpark_diverse,covernet,r2p2,sophie,tnt}. However, intermediate logic in DNN models is often not interpretable, and thus human designers have limited space to intervene with a prediction instance. For example, most prediction models do not allow to explicitly condition on a particular set of lanes on the road, while such decisions implicitly stem from random input noise. This is a practical drawback when the models are to be used in self-driving robots, causing them to sample inefficiently many trajectories to cover all plausible modes in the future.

In this paper, we propose a new trajectory prediction method, referred to as the {\it lane-aware prediction} (LaPred) model, which uses the instance-level lane information extracted from semantic maps. LaPred aims to predict the future trajectory of a target agent using a novel {\it per-lane joint feature}, which explicitly captures the complicated relation between a lane, a target agent, and a nearby agent at their instance level. Our per-lane joint feature explicitly aggregates the physical environment and the agent interaction at the instance-level, and it is useful to generate diverse trajectory samples from the distinct modes of distribution of future trajectories. We also propose an auxiliary model that determines which lane should be attended to predict the most plausible future trajectory by imposing a {\it self-supervised learning loss} \cite{self_supervised}. This auxiliary model guides LaPred to attend to the lanes that the target agent desires to follow, further enhancing the prediction accuracy. Our LaPred composes the per-lane joint feature, which accounts for the interaction with the most influential agent selected for each lane. With lane information, we can use a simple rule of choosing important nearby agents without having to use the complex attention networks employed in \cite {desire, sophie}. This can be seen as an advantage of using domain knowledge (e.g., lane and traffic information) to find  better representation of interactions with other agents. 

We evaluate the performance of LaPred on the public {\it Argoverse} dataset \cite{argoverse} and {\it nuScenes} dataset \cite{nuscenes}, which provide the semantic map along with the trajectory data. Our experiments demonstrate that LaPred produces reasonable prediction results for a variety of complex traffic scenarios and achieves a significant performance gain over existing methods. 
 
Our contributions are summarized as follows;
\begin{itemize}
    \item We propose a trajectory prediction network that uses the instance-level lane entities to represent the complicated relation between the lane and agent trajectories.
    \item We show that our per-lane joint feature is capable of finding better representation in both static and dynamic environments than other raster image-based methods, consequently producing the trajectories reflecting the lane constraints well.
    \item We propose an auxiliary model that learns to attend the lane candidates critical for trajectory prediction via a self-supervised classification task. This allows our model to generate high-quality trajectories reflecting important modes associated with lanes.
    \item We achieve state-of-the-art performance for some categories of Argoverse and nuScene benchmarks.
\end{itemize}

\section{Related Works} \label{relatedworks}
\subsection{Interaction-Aware Prediction}
Numerous methods have been proposed to predict the future trajectories of dynamic agents accounting for interaction with the neighboring agents. 
Grid-based pooling methods \cite{slstm, trivedi_ConvSP, desire, datadriven_interaction, SS-LSTM,matf} arrange the embedding vectors obtained by encoding the other agents' past motions to a regular tensor structure.
While grid-based pooling methods provide well-visualized ways to model social interaction, they require hand-craft designs for the coordinate systems and tensor resolution.

Global pooling methods \cite{soft+attn,sgan,shpark_diverse, sophie, mfp} treat embedding vectors without any spatial arrangements. Instead, they directly construct the embedding vectors through global pooling \cite{sgan} or neural attention architectures \cite{shpark_diverse,sophie, mfp}. Since there are no spatial constraints imposed by agent layout, these methods can process the interaction among an arbitrary number of agents without the race condition. Recently, spatiotemporal graph-based methods \cite{trajectron,bigat,trajectronpp} were employed to extend global pooling methods. Trajectron \cite{trajectron, trajectronpp} uses this graph structure to encode the dynamic influence of agents over time, and Social-BiGAT \cite{bigat} applies the attention mechanism to the graph to focus on the important parts of the interactions.

\subsection{Scene Context-Aware Prediction}
The majority of schemes generate 2D scene images using camera or LiDAR sensors. CAR-Net \cite{carnet} and SoPhie \cite{sophie} apply spatial attention to focus the salient regions relevant to trajectory prediction. MATF \cite{matf} constructs a tensor structure capturing both interactions between agents and scene contexts while retaining the spatial relationships. R2P2 \cite{r2p2} learns a policy model obtained by minimizing the symmetric cross-entropy, given the scene constraints.

The scene context can also be obtained from the information embedded in semantic maps. In \cite{multipath, mtp, tpnet, trivedi_mha, covernet, trajectronpp, mapadaptive}, 2D scene images were constructed by rasterizing the semantic map. Different types of information in the map can be encoded in each channel of an image. However, since trajectory data tend to be represented by the sequence of points in spatial coordinates, it is not trivial to reason the relationship between the trajectories and scene context (e.g., lanes) drawn on a 2D image. Hence, in \cite{vectornet, wimp, lanegcn, mtp_la, laneattention, tnt}, the instance-level representation of the scene context is used and described in the same coordinate domain as the trajectory.
Vectornet \cite{vectornet}, WIMP \cite{wimp}, LaneGCN \cite{lanegcn}, LA \cite{laneattention}, and TNT \cite{tnt} obtain the scene context from graph structures constructed from the instance-level semantic map information.
MTPLA \cite{mtp_la} estimates the most correlated lane using instance-level lane information.

While our method shares a similar idea to model instance-level information, it differs from previous methods in treating each lane and their associated agents in a single joint representation, leading to simpler model architecture while achieving empirical gains in the prediction performance. Furthermore, our method uses the auxiliary reference lane identification task to attend lanes that the target agent tries to follow.

\subsection{Generating diverse trajectory samples}
In practical applications, the models are often required to generate multi-modal trajectories with their likelihoods.
Generative models, including GAN and VAE, have been used to generate realistic trajectory samples \cite{sgan, desire, sophie}. These methods require large sample sets to achieve acceptable performance because the sample set should cover all trajectories with low probability but high importance in a traffic context. As remedies, DSF \cite{yuan2019diverse} proposes a diversity sampling function based on a determinantal point process. Diversity \cite{diversity} uses a latent space that controls the generation of semantically discrete trajectories, Multi-Path \cite{multipath} uses a fixed set of trajectory anchors that capture a driver's intentions. CoverNet \cite{covernet} formulates trajectory prediction as classification over the set of possible physically feasible trajectories. MTP \cite{mtp} proposes a multi-modal optimization function to generate multiple hypotheses. TNT \cite{tnt} and GoalNet \cite{mapadaptive} propose multi-modal generators to select targets (goals in \cite{mapadaptive}) among suggested candidates and generate trajectories based on the targets.

We conjecture that instance-level lanes are closely related to the modes of the trajectory distribution such that the lane-aware prediction eases the difficulty of finding the meaningful modes. We find that lane information offers a strong prior on the semantic behavior of a driver, and guides finding physically feasible modes in a traffic environment.

\section{Proposed Lane-Aware Multi-Modal Trajectory Prediction Method} \label{proposedmethod}
In this section, we present the details of the proposed LaPred method.

\subsection{Problem Formulation}
Suppose that the $N$ lane instances (called {\it lane candidates}) are found in the neighborhood of the target agent from the map where $N$ can vary with time.  We aim to predict the future trajectory sequence of the target agent based on its past trajectory sequence, the lane candidates,  and the nearby agents.
Each lane candidate is represented by the sequence of coordinates that are equally spaced and have the same length. 
For each lane candidate, the trajectory of a nearby agent that would interact the most with the target agent is identified by the preprocessing step described in subsection \ref{sec:preprocessing}.
Here are the notations used for our derivation;
\begin{itemize}
    \item $\mathbf{V}^{(p)} = \left\{V^{(p)}_{l-\tau}, V^{(p)}_{l-\tau+1}, ..., V^{(p)}_{l}\right\}$: the sequence of the past trajectory of the target agent over $(\tau+1)$ time steps where $l$ denotes the time index for the present, and $V^{(p)}_{l-i}$ is the coordinate of the target agent's centroid relative to $V^{(p)}_l=(0,0)$.
    \item $\mathbf{V}^{(f)} = \left\{V^{(f)}_{l+1},V^{(f)}_{l+2},...,V^{(f)}_{l+h}\right\}$: the sequence of the future trajectory of the target agent over $h$ time steps.
    \item $\mathbf{L}^{n}= \left\{L_{1}^{n},L_{2}^{n},...,L_{M}^{n}\right\}$: the sequence of $M$ equally spaced coordinate points on the center lane of the $n$th lane candidate. 
    \item $\mathbf{V}^{n} =\left\{V_{l-\tau}^{n},V_{l-\tau+1}^{n},...,V_{l}^{n}\right\}$: the sequence of the past trajectory of a single nearby agent selected for the $n$th lane candidate.
\end{itemize}
We term the lane candidate that the target agent tries to follow a {\it reference lane}.
We assume that the reference lane always exists among the set of $N$ lane candidates $\mathbf{L}^1,...,\mathbf{L}^{N}$.

\begin{figure*}[tbh]
        \centering
        \includegraphics[width=1.0\textwidth]{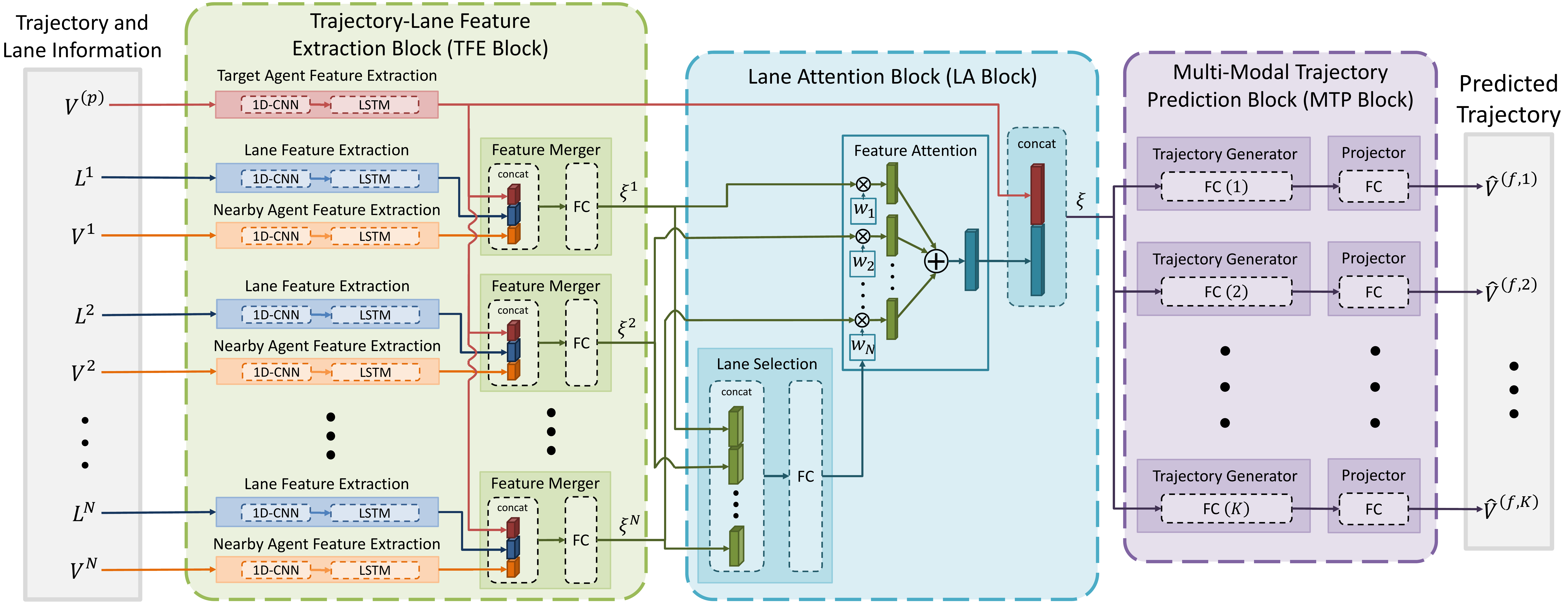}
     \caption {{\bf Overall structure of LaPred Network:} The past trajectories of the target and nearby agents and the lane candidates are fed into the TFE block. The TFE block generates the trajectory-lane features for each lane candidate. The LA block produces the joint representation of the observations via weighted aggregation of the trajectory-lane features. Finally, the $K$ multiple trajectory samples are generated in MTP block.  }
    \label{fig:all_system}
    %\vspace{-5mm}
\end{figure*}

The trajectory prediction task can be formulated as finding the posterior distribution of the future trajectory given all available observations, $p\left(\mathbf{V}^{(f)}|\mathbf{V}^{(p)},\mathbf{L}^{1:N}, \mathbf{V}^{1:N}\right)$ where $\mathbf{L}^{1:N}=\left\{\mathbf{L}^{1},...,\mathbf{L}^{N} \right\}$ and $\mathbf{V}^{1:N}=\left\{\mathbf{V}^{1},...,\mathbf{V}^{N} \right\}$.
Let $E_i$ be the event that the $i$th lane candidate $\mathbf{L}^{i}$ becomes the reference lane. Then, the conditional distribution $p(\mathbf{V}^{(f)}|\mathbf{V}^{(p)},\mathbf{L}^{1:N}, \mathbf{V}^{1:N})$  can be expressed as 
\begin{align}
&p\left(\mathbf{V}^{(f)}|\mathbf{V}^{\left(p\right)},\mathbf{L}^{1:N}, \mathbf{V}^{1:N}\right) \nonumber \\ 
&=\sum_{i=1}^{N} p\left( \mathbf{V}^{(f)},E_i\Big|\mathbf{V}^{(p)},\mathbf{L}^{1:N}, \mathbf{V}^{1:N} \right) \nonumber \\
&=\sum_{i=1}^{N} p\left( \mathbf{V}^{(f)}\Big|E_i,\mathbf{V}^{(p)},\mathbf{L}^{1:N}, \mathbf{V}^{1:N} \right) \nonumber \\
&\quad \quad p\left( E_i\Big|\mathbf{V}^{(p)}, \mathbf{L}^{1:N}, \mathbf{V}^{1:N} \right). \label{eq:obj_xi}
\end{align}
In (\ref{eq:obj_xi}), we assume that no two or more lane candidates can be the reference lane and the reference lane always exists among lane candidates, i.e.,  $p(E_i \cap E_j) = 0$ and $p(\cup_{i=1}^{N}E_i)=1$. The first term in (\ref{eq:obj_xi}) indicates the distribution of the future trajectory $\mathbf{V}^{(f)}$ under the condition that the $i$th lane candidate is identified as a reference lane. Conditioned on $E_i$, the target agent tries to follow the $i$th lane candidate, interacting the nearby agent on the $i$th lane. Hence, we assume that conditioned on $E_i$, the future trajectory depends solely on $\mathbf{L}^{i}$ and $\mathbf{V}^{i}$, i.e., 
\begin{align}
        &p\left(\mathbf{V}^{(f)}\Big|E_i,\mathbf{V}^{(p)},\mathbf{L}^{1:N}, \mathbf{V}^{1:N} \right) \nonumber \\
        &= p\left(\mathbf{V}^{(f)}\Big|E_i, \mathbf{V}^{(p)},\mathbf{L}^{i}, \mathbf{V}^{i} \right) \label{eq:agg}   
\end{align}
Finally, the conditional distribution in (\ref{eq:obj_xi}) becomes
\begin{align}
&p\left(\mathbf{V}^{(f)}\Big|\mathbf{V}^{\left(p\right)},\mathbf{L}^{1:N}, \mathbf{V}^{1:N}\right) \nonumber \\ 
&=\sum_{i=1}^{N} p\left(\mathbf{V}^{(f)}\Big|E_i, \mathbf{V}^{(p)},\mathbf{L}^{i}, \mathbf{V}^{i} \right) \nonumber \\
& \quad \quad p\left( E_i \Big|\mathbf{V}^{\left(p\right)},\mathbf{L}^{1:N}, \mathbf{V}^{1:N}\right) . \label{eq:obj_xi2}
\end{align}
This shows that the prediction of the future trajectory can be achieved by aggregating the trajectory predicted under condition $E_i$, weighted with the probability of $E_i$ given all observations.

The encoder-decoder structure has been shown to be effective in predicting target sequences of different lengths based on the source sequence \cite{shpark}. Thus, we adopt the encoder-decoder structure to obtain the conditional distribution in (\ref{eq:obj_xi2}).
Let $\xi^{i}$ be the abstract representation of $ \{\mathbf{V}^{(p)},\mathbf{L}^{i}, \mathbf{V}^{i} \}$ associated with the $i$th lane candidate.
  Inspired by the equation (\ref{eq:obj_xi2}), the encoder generates the context vector $\xi$ as
\begin{align}
    \xi &=\sum_{i=1}^{N}  \xi^i  p\left( E_i \Big|\mathbf{V}^{\left(p\right)},\mathbf{L}^{1:N}, \mathbf{V}^{1:N} \right) \\
    &\approx \sum_{i=1}^{N}  \xi^i  p\left( E_i \Big| \xi^{1:N} \right).
\end{align}
Then, the decoder learns the predictive distribution $p\left(\mathbf{V}^{f}\big|\xi\right)$, from which the multiple trajectory samples are generated. 
Motivated by the derivation above, the trajectory prediction of LaPred consists of the following steps
\begin{itemize}
    \item {\bf Trajectory-lane feature extraction:} extract the trajectory-lane features  $\xi^{i}$ from $\mathbf{V}^{(p)},\mathbf{L}^{i}, \mathbf{V}^{i}$. 
    \item {\bf Reference lane identification:} find the probability  of $E_i$   given
 the set of the trajectory-lane features $\xi^{1:N}$, i.e., $p\left( E_i \big|\xi^{1:N} \right)$. The probability $p\left( E_i \big|\xi^{1:N} \right)$ is considered as the attention weight given to the $i$th lane candidate. 
        \item {\bf Weighted feature aggregation:} compute the weighted average of the trajectory-lane features according to $\xi = \sum_{i=1}^{N} \xi^{i} p\left( E_i \Big|\xi^{1:N} \right)$.
        \item {\bf Trajectory decoding:} generate the future trajectories according to $p\left(\mathbf{V}^{f}\big|\xi\right)$. 
        \end{itemize}

\subsection{Structure of LaPred Network}
In this section, we describe the detailed structure of the proposed LaPred network.

\subsubsection{Overall System}
Fig. \ref{fig:all_system} depicts the overall structure of LaPred network. First, the preprocessing block extracts the lane candidates $\mathbf{L}^{1:N}$ from the map. The trajectory sequences are collected for the $N$ neighboring agents $\mathbf{V}^{1:N}$ selected for each lane candidate.    
The trajectory-lane feature extraction (TFE) block extracts the features $\xi^{i}$ from $\mathbf{V}^{(p)},\mathbf{L}^{i}, \mathbf{V}^{i}$. Note that the parameters of the TFE block are shared over the lane candidates. Based on $N$ trajectory-lane features $\mathbf{\xi}^{1:N}$,  the lane attention (LA) block calculates the attention weights given by the probability that the $i$th candidate lane is identified as a reference lane. The trajectory-lane features $\xi^{1:N}$ are weighted by the attention weights and combined to produce the joint representation, ${\xi}$. Finally, the multi-modal trajectory prediction (MTP) block produces $K$  multi-modal trajectories based on the aggregated features $\xi$.

\subsubsection{Preprocessing} \label{sec:preprocessing}
In the preprocessing step, the $N$ lane candidates are identified based on the distance from the current position of the target agent. First, we search for the lane segments (a set of coordinates comprising the center of a lane segment) within the search radius (e.g., $10$ meters) from the centroid of the target agent. Then, we extend the lane segments by attaching the preceding and succeeding lane segments based on lane connectivity information in the map until the length of the extended lane reaches a predefined value. When there are more than $N$ connected lane segments, only $N$ lane instances are selected based on Euclidean distance from the current location of the target agent. The selected $N$ lane instances become {\it lane candidates}. The set of coordinates for these $N$ lane candidates are resampled such that any two adjacent coordinate points have equal distance.

The lane candidate closest to the future trajectory of the target agent is labeled as a reference lane.
The distance between the lane candidate and the trajectory is measured by 
\begin{align}
    &{\mathcal{D}\left ( \mathbf{V}^{(f)},\mathbf{L}^n \right )}=\sum^{h}_{i=1} \eta(i)  \min_{m\in \left \{ 1,...,M \right \}} \left \| V_{l+i}^{\left ( f \right )} - L^n_{m} \right \|, \label{eq:dist}
\end{align}
where $\eta(i)$ is the scaling weight applied for different time steps. The reference lane is decided such that higher weight is applied for the lane points on farther horizons e.g., $\eta(i)=i$. Note that the lane candidates are autonomously labeled  using the aforementioned criterion without manual labor.

The preprocessing step also searches for the nearby agents, which are like to have the most influence on the trajectory of the target agent. It selects only one nearby agent for each lane candidate. Specifically, all nearby agents are identified within the fixed range from the center point of each lane candidate, and the nearest one in front of the target agent is selected as the most influential agent.

\subsubsection{Trajectory-Lane Feature Extraction}
The TFE block extracts the joint trajectory-lane features $\mathbf{\xi}^{i}$ from the observation $\{\mathbf{V}^{(p)},\mathbf{L}^{i}, \mathbf{V}^{i}\}$ for the $i$th lane candidate.
The observations $\mathbf{V}^{(p)},\mathbf{L}^{i}$, and $\mathbf{V}^{i}$ are separately encoded by the one-dimensional (1D)-CNN followed by the long short term memory (LSTM) model
\begin{align}
    \xi_{V_p} &= {\rm LSTM}(\mbox{1D-CNN}(\mathbf{V}^{(p)})) \\
    \xi_{L_i} &= {\rm LSTM}(\mbox{1D-CNN}(\mathbf{L}^i))\\
    \xi_{V_i} &= {\rm LSTM}(\mbox{1D-CNN}(\mathbf{V}^i)).
\end{align}
These CNN-LSTM encoding networks have different weights for each of $\mathbf{V}^{(p)},\mathbf{L}^{i}$ and $\mathbf{V}^{i}$. The weights of the entire TFE block are shared over the lane candidates. The features $\xi_{V_p},\xi_{L_i}$ and $\xi_{V_i}$  are concatenated and fed through the fully connected (Fc) layers to produce the joint features  $\mathbf{\xi}^{i}$ for the $i$th lane candidate. 

\subsubsection{Lane Attention}
The LA block produces the joint representation $\mathbf{\xi}$ of the given conditions via a weighted combination of the $N$ trajectory-lane features $\mathbf{\xi}^{1:N}$.
As mentioned, the attention weight $w_i$ for the $i$th lane candidate is obtained from $p\left( E_i \big|\xi^{1:N} \right)$. The attention weight $w_i$ is obtained by concatenating the $N$ trajectory-lane features $\mathbf{\xi}^{1:N}$ and applying the Fc layers followed by the soft-max function. We consider the fixed number of lane candidates $N$, but sometimes, the number of lane candidates found in the preprocessing block can be less than $N$. In this case, zero vectors are fed into the model as an input. To allow the model to better attend the lane candidates important for trajectory prediction, we impose the additional reference lane identification task on top of the prediction task. The LA model performs the task of selecting the reference lane among $N$ lane candidates supervised by the auto-labeled data. This enables the regularization of our model through the auxiliary self-supervised learning loss \cite{self_supervised}.
Finally, the trajectory-lane features $\mathbf{\xi}^{1:N}$ are combined using the attention weight $w_i$ as
    $\mathbf{\xi} = \sum_{i=1}^{N} w_i \mathbf{\xi}^{i}$.
 As an approximation to our weighted feature aggregation, we can also consider the hard selection, which forces $w_i=1$ for the lane candidate with the highest attention weight and sets $w_i=0$ for the rest.  In the experiment section, we evaluate the benefit of the soft selection over the hard selection.

\subsubsection{Multi-Modal Trajectory Prediction}
We first construct the complete joint representation by concatenating the features $\xi_{V_p}$ to the output $\xi$ of the LA block through the skip connection.
The MTP block generates $K$ hypotheses of the future trajectory of the target agent %($\hat{V}^{(f),1:K}$) 
based on the joint representation $\{ \mathbf{\xi}, \xi_{V_p}\}$. To generate the multiple trajectory samples, we employ the multi-modal generator model proposed in \cite{mtp}. The $K$ trajectory samples $\hat{\mathbf{V}}^{(f,1)},...,\hat{\mathbf{V}}^{(f,K)}$ are generated using the $K$ sample generator models. Each generator model is divided into two parts; 1) the Fc layers not shared across the $K$ models and 2) the subsequent Fc layers shared.  We use the shared Fc layers to alleviate the over-fitting problem that arises by using only a partition of the training data for each generator model. While only one of the $K$ non-shared Fc layers is updated for the given input, the shared Fc layers are all updated for the input.  

\subsection{Training Details}
The loss function $L_{total}$ used to train the proposed LaPred model is given by
\begin{align}
L_{total} =  \alpha L_{pred} + \left ( 1 - \alpha  \right ) L_{cls},
\label{eq:loss_total}
\end{align}
where $L_{pred}$ denotes the mean absolute error loss, and $L_{cls}$ denotes the cross-entropy loss for selecting the reference lane from the lane candidates. Since the MTP block produces the $K$ outputs simultaneously, $L_{pred}$ is given by 
\begin{align}
    L_{pred} = \sum_{t\in Batch} \min_{k\in \left \{ 1,\cdots ,K \right \}} L_{pred}^{t,k},
    \label{eq:loss_pred}
\end{align}
where $L_{pred}^{t,k}$ denotes the loss function for the $k$th generator model evaluated for the $t$th training sample.
The loss $L^{t,k}_{pred}$ is expressed as
\begin{align}
    L_{pred}^{t,k} = \beta L_{pos}^{t,k} + \left ( 1 - \beta  \right ) L_{\rm lane-off}^{t,k},
    \label{eq:loss_k}
\end{align}
where $L_{pos}^{t,k}$ is a smooth $L_1$ loss between $\hat{\mathbf{V}}^{(f,k)}$ and $\mathbf{V}^{(f)}$.
To reflect the tendency of the target agent stick close to the reference lane in the future, we devise a new loss function $L_{\rm lane-off}$, which is expressed as
\begin{align}
L_{\rm lane-off}^{t,k} = \frac{1}{h} \sum^{h}_{i=1} l\left( \hat{V}_{l+i}^{\left( f,k \right)}, V_{l+i}^{\left( f,k \right)},  \mathbf{L}^{\left( ref \right)} \right),
\label{eq:loss_lo}
\end{align}
where % $l_{lane-off,i}^{k}$ is given by
$\mathbf{L}^{\left( ref \right)}$ is the lane instance selected as a reference lane and
\begin{align}
& l \left(\hat{V}, V,  \mathbf{L} \right) = \left\{
\begin{array}{ccc}
 \delta  ( \hat{V}, \mathbf{L}  )
%- \delta  ( V , \mathbf{L} )    
& \text{ if } \delta ( \hat{V}, \mathbf{L}  )
> \delta \left ( V , \mathbf{L} \right ) \\ 
  0 & \text{ otherwise, }
 \end{array}
 \right. \nonumber
\end{align}
where $\delta \left( V,\mathbf{L} \right)$ denotes the distance from the point $V$ to the lane $\mathbf{L}$. This loss function encourages the model to reduce the distance from the lane whenever the prediction deviates from a lane farther than the ground truth.

\section{Experiments} \label{experiments}
In this section, we evaluate the performance of the proposed LaPred method.

\begin{table*}[tbh]
\centering
\begin{tabular}{c|rrrrrr}
\hline
Methods      & \multicolumn{1}{c|}{A} & \multicolumn{1}{c|}{B} & \multicolumn{1}{c|}{C} & \multicolumn{1}{c|}{D} & \multicolumn{1}{c|}{E}  \\ \hline
Lane Information & \multicolumn{1}{c}{} & \multicolumn{1}{c}{$\surd$} & \multicolumn{1}{c}{$\surd$}  & \multicolumn{1}{c}{$\surd$} & \multicolumn{1}{c}{$\surd$} \\ \cline{1-1}
%{$\xi_{total}$ from ground-truth reference lane} & \multicolumn{1}{c}{} & \multicolumn{1}{c}{} & \multicolumn{1}{c}{$\surd$} & \multicolumn{1}{c}{} & \multicolumn{1}{c}{} &   \\ \cline{1-1}
{$L_{\rm lane-off}$}     & \multicolumn{1}{c}{} & \multicolumn{1}{c}{} & \multicolumn{1}{c}{$\surd$} & \multicolumn{1}{c}{$\surd$} & \multicolumn{1}{c}{$\surd$}  \\ \cline{1-1}
{Nearby Agents} & \multicolumn{1}{c}{} & \multicolumn{1}{c}{} & \multicolumn{1}{c}{} & \multicolumn{1}{c}{$\surd$} & \multicolumn{1}{c}{$\surd$}  \\ \cline{1-1}
%{Hard selected $\xi_{total}$} & \multicolumn{1}{c}{} & \multicolumn{1}{c}{$\surd$} & \multicolumn{1}{c}{$\surd$} & \multicolumn{1}{c}{$\surd$} &   \\ \cline{1-1}
{Soft selected $\xi_{agg}$} & \multicolumn{1}{c}{} & \multicolumn{1}{c}{} & \multicolumn{1}{c}{} & \multicolumn{1}{c}{} & \multicolumn{1}{c}{$\surd$} \\ \hline
$ADE_1$         & $4.77$  & $4.04_{\downarrow0.73}$ & $4.00_{\downarrow0.04}$ & $3.82_{\downarrow0.18}$ & $3.51_{\downarrow0.31}$ \\
$FDE_1$         & $11.10$ & $9.30_{\downarrow1.80}$ & $9.23_{\downarrow0.07}$ & $8.90_{\downarrow0.33}$ & $8.12_{\downarrow0.78}$ \\
%$ADE_3$ & $2.47$  & $2.27_{\downarrow0.20}$ & $2.25_{\downarrow0.02}$ & $2.23_{\downarrow0.02}$ & $1.96_{\downarrow0.27}$ \\
%$FDE_3$ & $5.35$  & $4.91_{\downarrow0.44}$ & $4.89_{\downarrow0.02}$ & $4.79_{\downarrow0.10}$ & $4.39_{\downarrow0.40}$ \\
$ADE_5$ & $2.48$  & $2.06_{\downarrow0.42}$ & $1.77_{\downarrow0.29}$ & $1.72_{\downarrow0.05}$ & $1.53_{\downarrow0.19}$ \\
$FDE_5$ & $5.33$  & $4.37_{\downarrow0.96}$ & $3.62_{\downarrow0.75}$ & $3.54_{\downarrow0.08}$ & $3.37_{\downarrow0.17}$ \\ \hline
\end{tabular}
\caption{Ablation study conducted on nuScenes validation set}
\label{table:ablation}
\end{table*}

\begin{table*}[tbh]
\centering
\begin{tabular}{c|cccccccc}
Network & \multicolumn{1}{c}{$ADE_1$} & \multicolumn{1}{c}{$FDE_1$} & \multicolumn{1}{c}{$ADE_5$} & \multicolumn{1}{c}{$FDE_5$} & \multicolumn{1}{c}{$ADE_{10}$} & \multicolumn{1}{c}{$FDE_{10}$} & \multicolumn{1}{c}{$ADE_{15}$} & \multicolumn{1}{c}{$FDE_{15}$} \\ \hline
MTP \cite{mtp}& 4.42 & 10.36 & 2.22 & 4.83 & 1.74 & 3.54 & 1.55 & 3.05 \\
MultiPath \cite{multipath}& 4.43 & 10.16 & 1.78 & 3.62 & 1.55 & 2.93 & 1.52 & 2.89 \\
CoverNet \cite{covernet}& 3.87 & 9.26  & 1.96 &  -   & 1.48 &  -   &  -   &  -  \\
Trajectron++ \cite{trajectronpp}&  -   & 9.52  & 1.88 &  -   & 1.51 &  -   &  -   &  -   \\
MHA-JAM \cite{trivedi_mha}&  3.69  &  8.57  &  1.81  &  3.72   &  1.24   &  \textbf{2.21}  &  \textbf{1.03}  &  \textbf{1.7} \\ \hline
Ours         &  \textbf{3.51}  &  \textbf{8.12}  &  \textbf{1.53} &  \textbf{3.37}   &  \textbf{1.12}  &  2.39  &  1.10 &  2.34 
\end{tabular}
\caption{Performance of several prediction methods evaluated on nuScenes validation set}
\label{table:nuscenes}
\end{table*}

\begin{table*}[tbh]
\centering
\begin{tabular}{c|cccccccc}
Network & \multicolumn{1}{c}{$ADE_1$} & \multicolumn{1}{c}{$FDE_1$} & \multicolumn{1}{c}{$ADE_5$} & \multicolumn{1}{c}{$FDE_5$} & \multicolumn{1}{c}{$ADE_{6}$} & \multicolumn{1}{c}{$FDE_{6}$} & \multicolumn{1}{c}{$ADE_{12}$} & \multicolumn{1}{c}{$FDE_{12}$} \\ \hline
%MATF \cite{matf}&\textbf{1.35}&\textbf{2.50}& 1.29 & 2.38 & 1.28 & 2.37 & 1.26 & 2.31 \\
DESIRE\cite{desire}$^{*}$& 2.38 & 4.64 & 1.17 & 2.06 & 1.09 & 1.89 & 0.90 & 1.45 \\
R2P2\cite{r2p2}$^{*}$& 3.02 & 5.41 & 1.49 & 2.54 & 1.40 & 2.35 & 1.11 & 1.77 \\
VectorNet \cite{vectornet}& 1.66 & 3.67 &   -  &   -  & - & - & - & - \\
DiversityGAN \cite{diversity}&   -  &   -  & 1.33 & 2.72 &   -  &   -  & - & - \\
MFP \cite{mfp}&   -  &   -  &   -  &   -  & 1.40 & - &   -  & - \\
DATF\cite{shpark_diverse}$^{*}$& 2.04 & 3.69 &   0.98  &  1.65  & 0.92 & 1.52 & 0.73 & \textbf{1.12} \\
MTPLA \cite{mtp_la}&\textbf{1.46}&\textbf{3.27}&   -  &   -  & 1.05 & 2.06 & - & - \\ \hline
Ours & 1.48 & 3.29 &\textbf{0.76}&\textbf{1.55}& \textbf{0.71} & \textbf{1.44} & \textbf{0.60} & 1.15 \\
\multicolumn{9}{r}{$*$ indicates our own implementation.}
\end{tabular}
\vspace*{-4mm}
\caption{Performance of several prediction methods evaluated on Argoverse validation set}
\label{table:argoverse}
\end{table*}

\subsection{Dataset}
Two public datasets, nuScenes and Argoverse, are used to evaluate the performance of LaPred. These are the only datasets that provide the high definition (HD) map associated with the trajectory data. nuScenes dataset \cite{nuscenes} provides the log of the ego vehicle's state, the annotations of nearby agents' location, and HD-map data. The dataset provides 245,414 trajectory instances in 1,000 different scenes. The trajectory instances consist of the sequence of two dimensional (2D) coordinates for 8 seconds duration sampled at $2 Hz$. The trajectory prediction task defined in nuScenes benchmark is to predict a 6-second future trajectory based on a 2-second past trajectory for each target agent.

Argoverse forecasting dataset \cite{argoverse} is the dataset specialized for trajectory prediction tasks.
The dataset contains the trajectories of the target agent, those of nearby agents, and HD-map data.
A total of 324,557 scenarios are included, each of which is 5 seconds. The 2D coordinates comprising trajectories are sampled at $10 Hz$. The trajectory prediction task is defined as predicting a 3-second future trajectory based on a 2-second past trajectory.

\subsection{Experiment Results}
In this section, we evaluate the performance of LaPred. We employ two popularly used evaluation metrics, {\it average displacement error} (ADE) and {\it final displacement error} (FDE). Let $e^{k}_{i} = \norm{V^{(f)}_{l+i}-\hat{V}^{(f,k)}_{l+i}}_2$, then the two evaluation metrics
$ADE_{K}$ and $FDE_{K}$  are defined as
\begin{align}
{ADE_{K}} &= \E_{p} \underset{k \in \{1,...,K\}}{\min}\left(\frac{1}{h}\sum^{h}_{i=1}e^{k}_{i}\right) \\
{FDE_{K}} &= \E_{p}\underset{k\in \{1,...,K\}}{\min}e^{k}_h.
\end{align}

\subsubsection{Ablation Study}
\begin{figure*}[tbh]
    %\captionsetup[subfigure]{labelformat=empty}
    \captionsetup[subfigure]{}
    \centering
    %\subfloat[]{
    %\raisebox{1.5cm}{\rotatebox[origin=t]{90}{K=1}}}\hspace{0.1cm}
    \subfloat[]{
    \includegraphics[width=4cm,height=4cm]{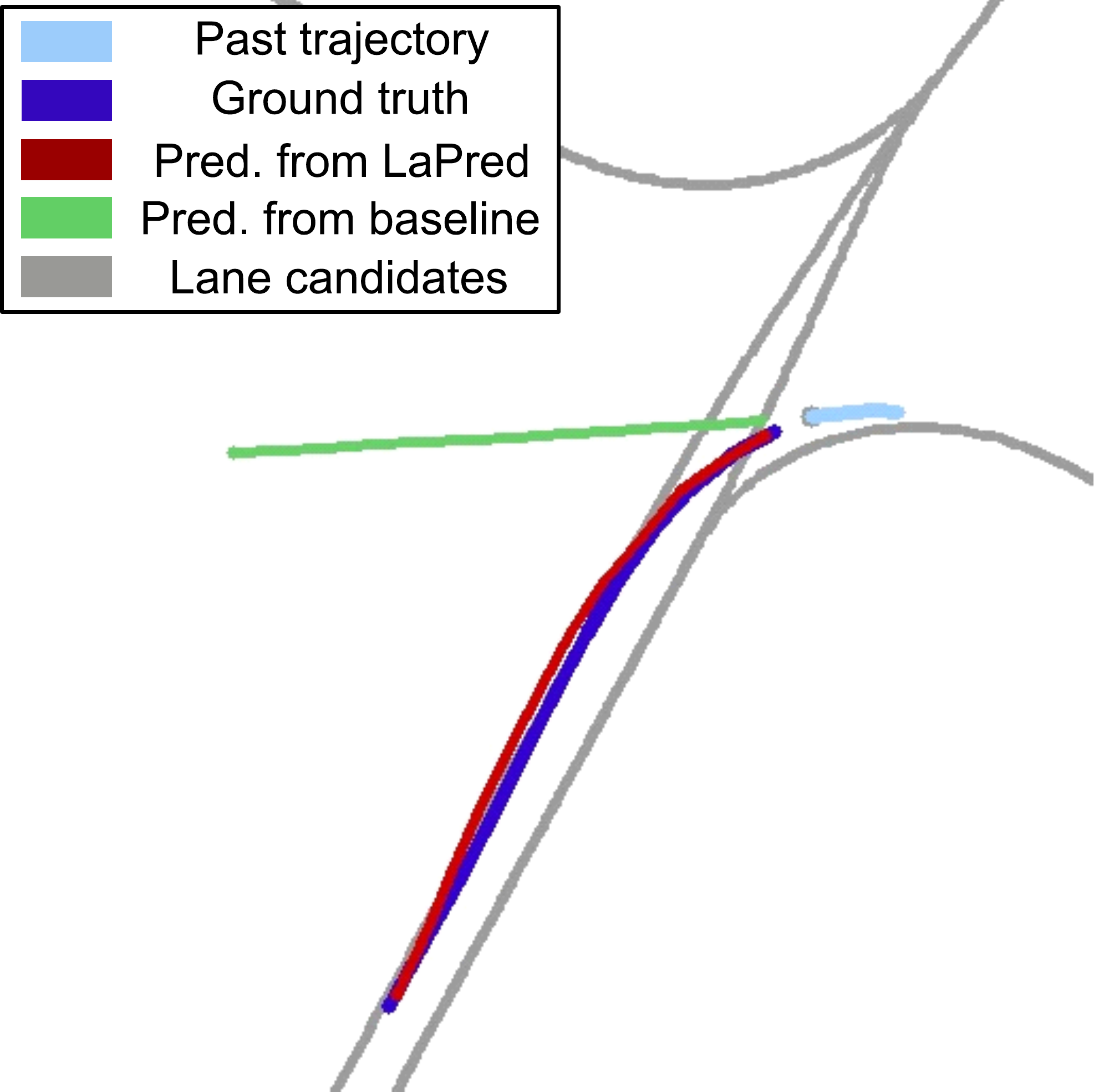}}\hspace{0cm}
    \subfloat[]{
    \includegraphics[width=4cm,height=4cm]{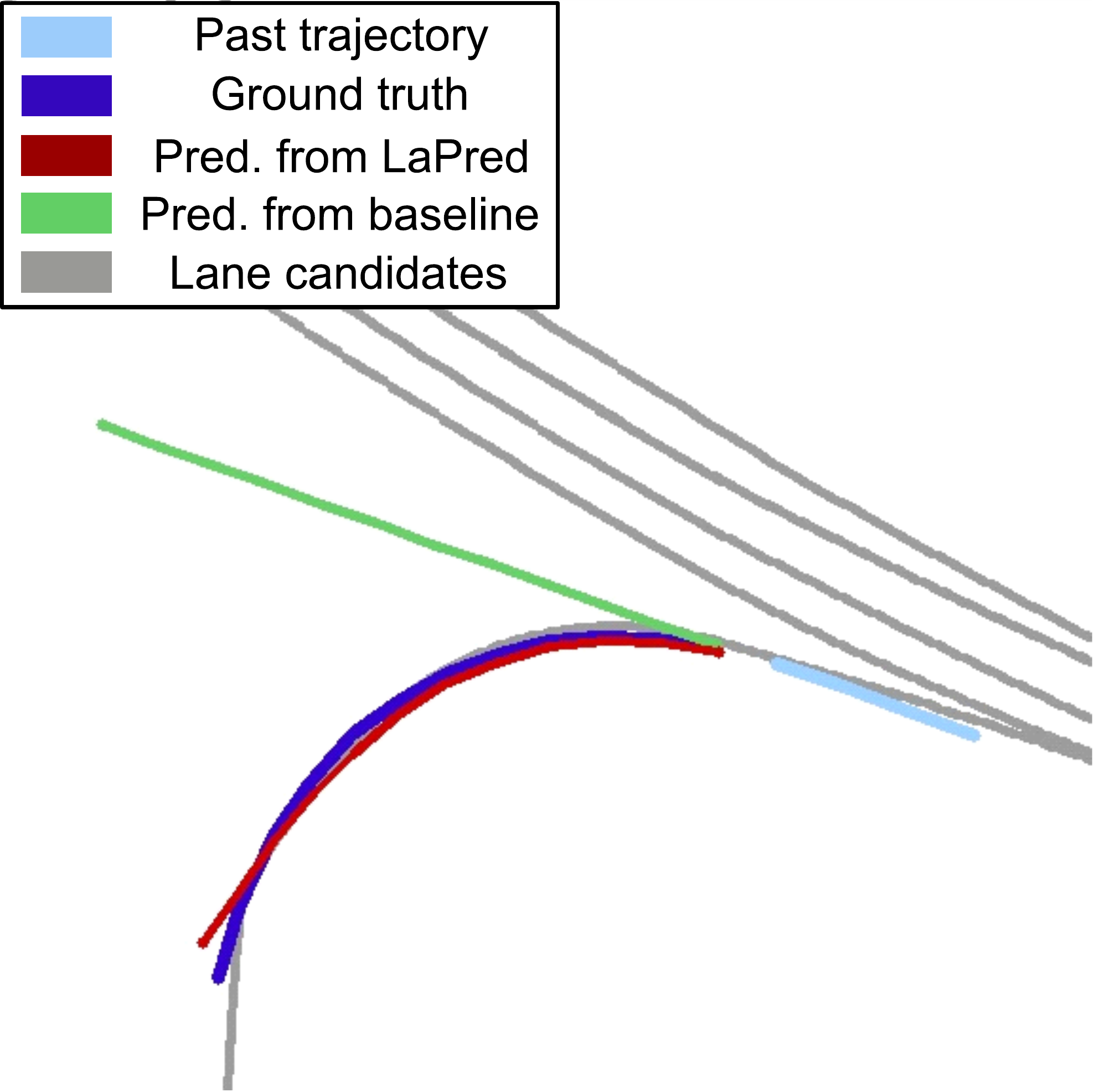}}\hspace{0cm}
    \subfloat[]{
    \includegraphics[width=4cm,height=4cm]{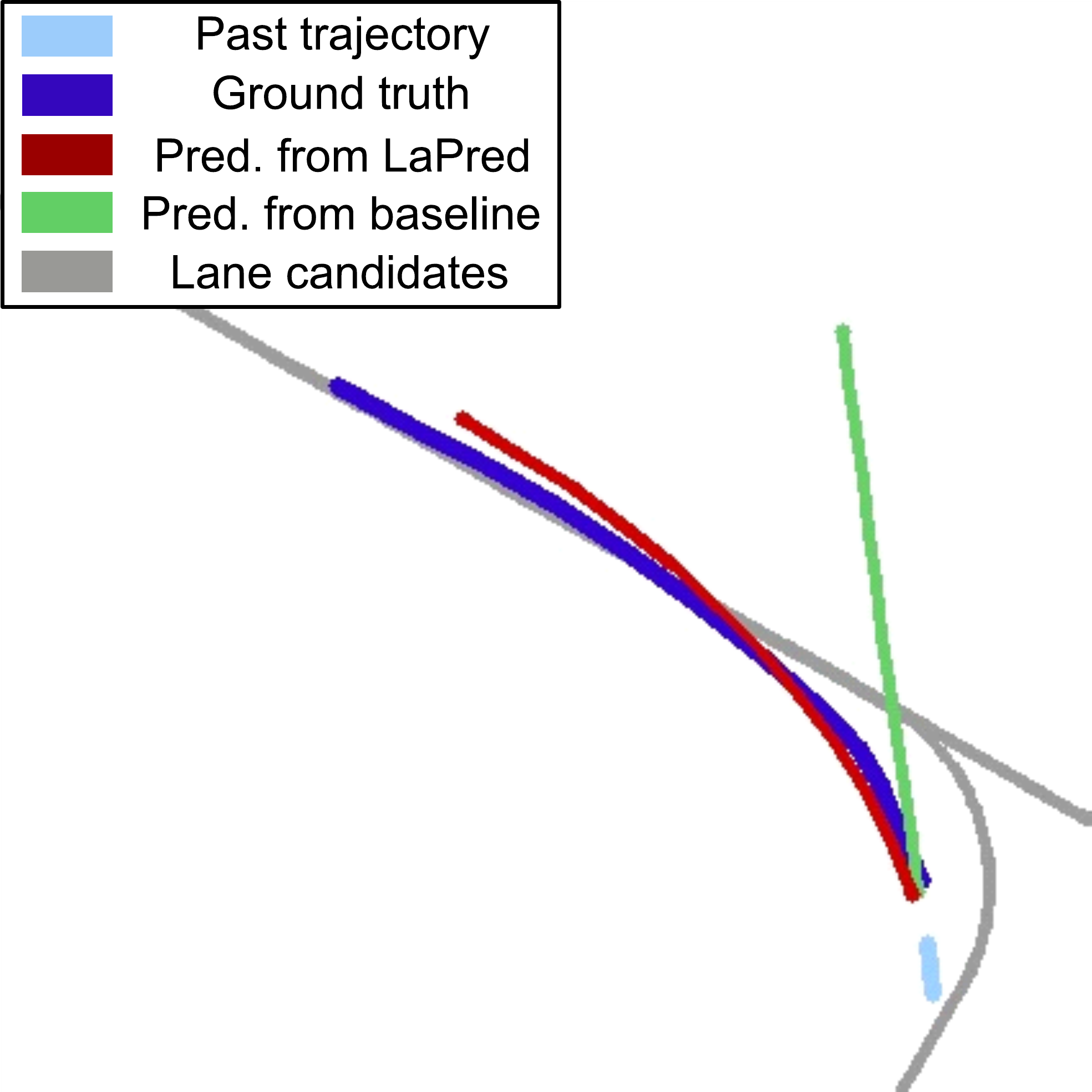}}\hspace{0cm}
    \subfloat[]{
    \includegraphics[width=4cm,height=4cm]{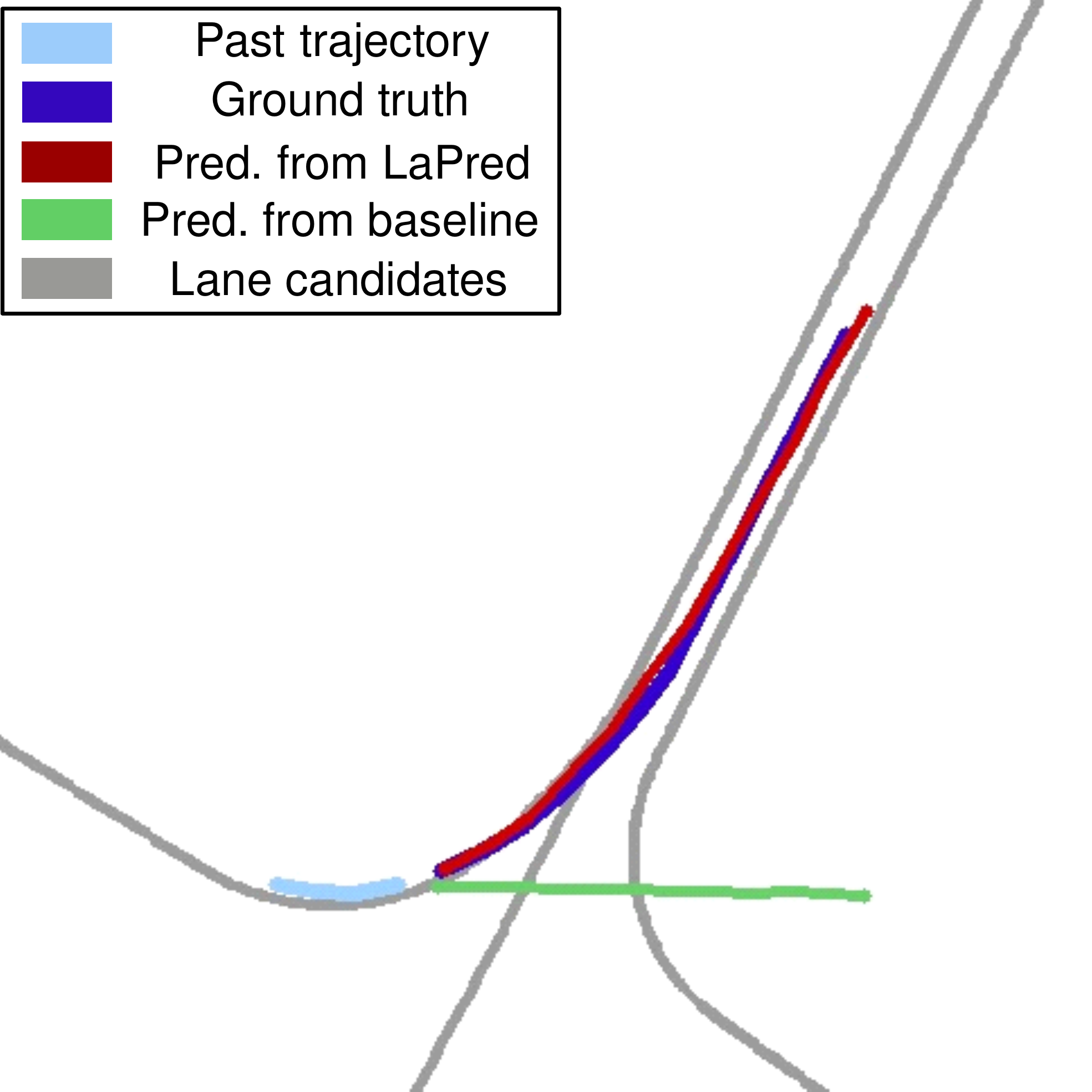}}
     %\bigskip
    %\captionsetup[subfigure]{labelformat=empty}
    \captionsetup[subfigure]{}
    \centering
    %\subfloat[]{
    %\raisebox{1.5cm}{\rotatebox[origin=t]{90}{K=5}}}\hspace{0.1cm}
    \subfloat[]{
    \includegraphics[width=4cm,height=4cm]{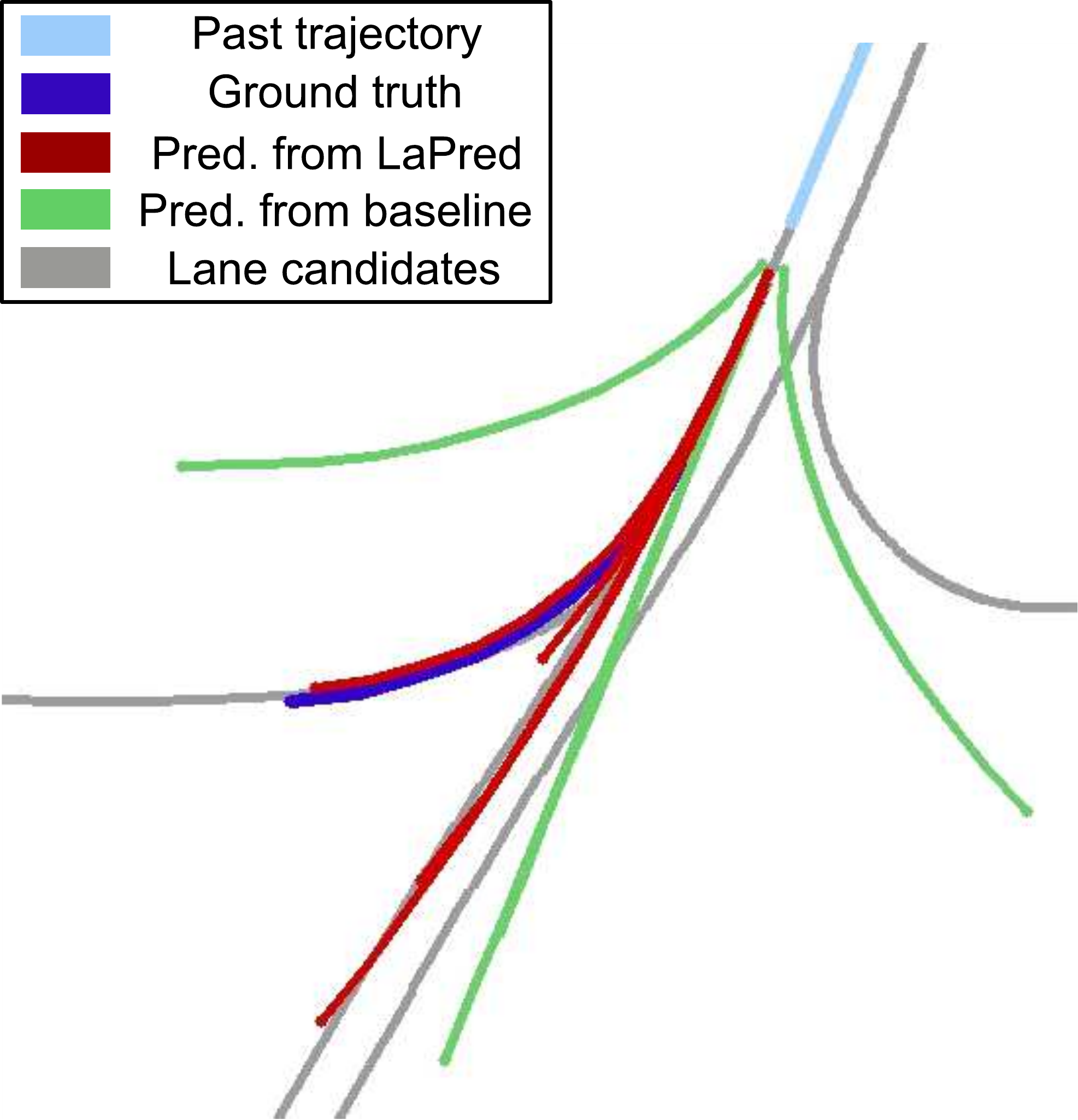}}\hspace{0cm}
    \subfloat[]{
    \includegraphics[width=4cm,height=4cm]{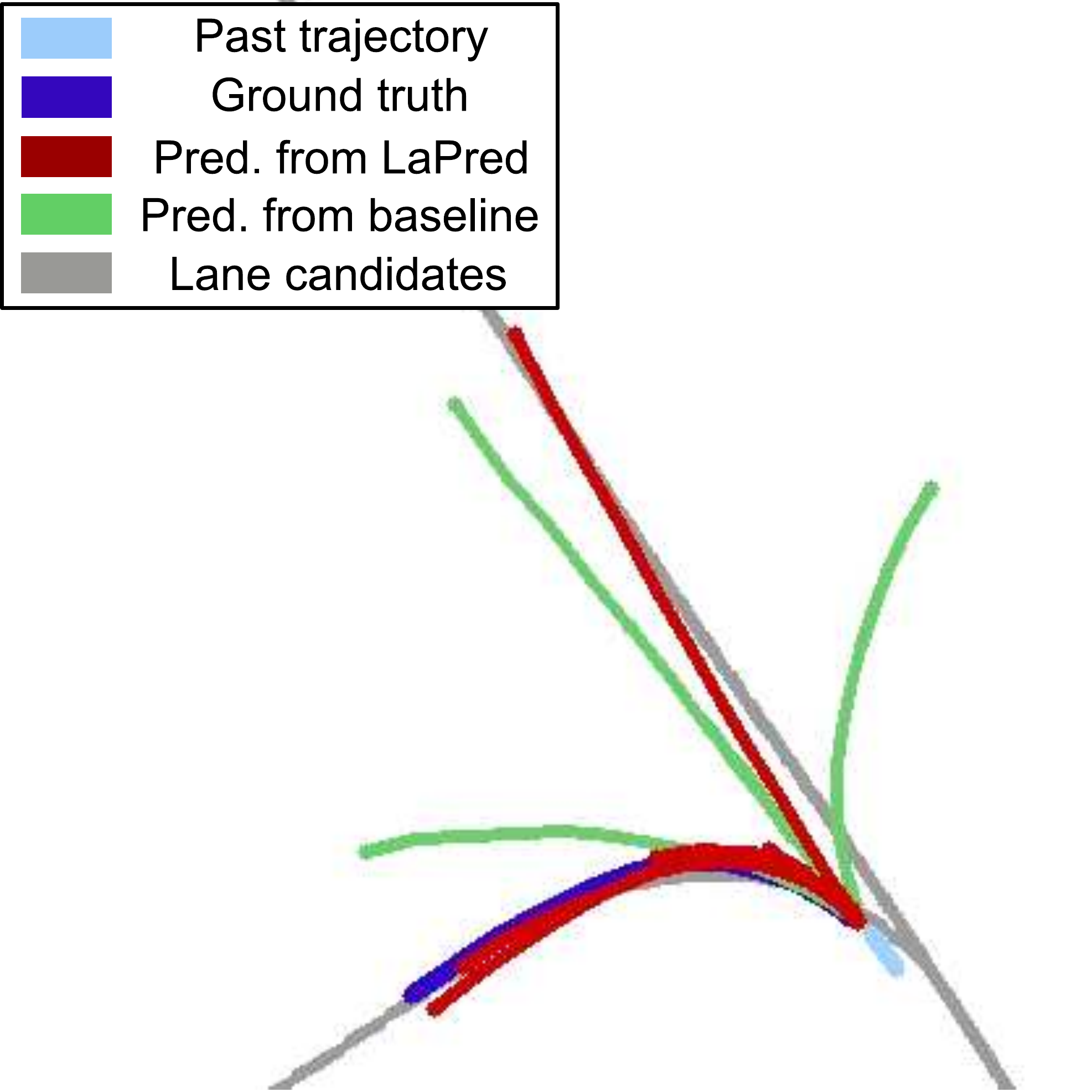}}\hspace{0cm}
    \subfloat[]{
    \includegraphics[width=4cm,height=4cm]{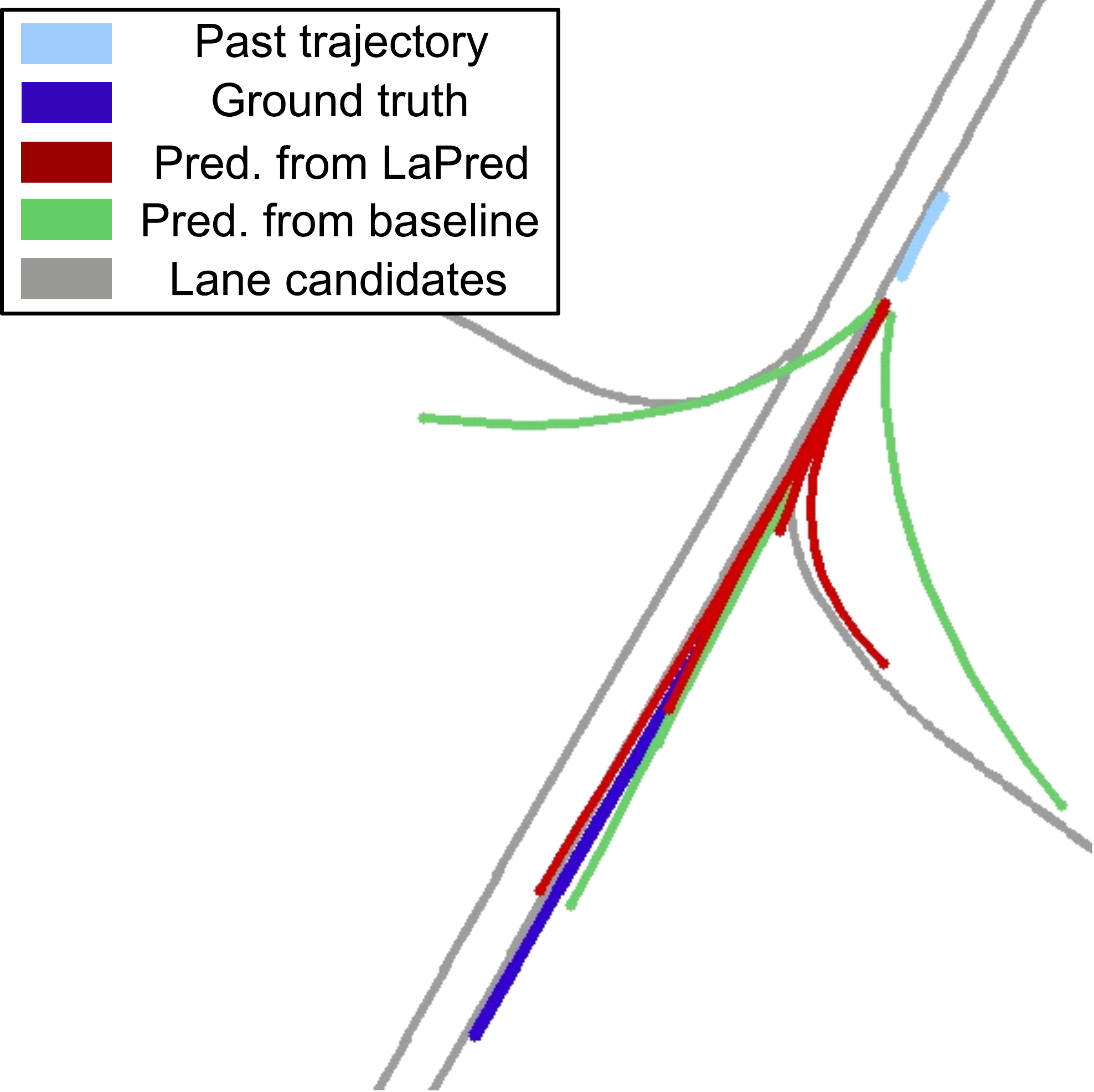}}\hspace{0cm}
    \subfloat[]{
    \includegraphics[width=4cm,height=4cm]{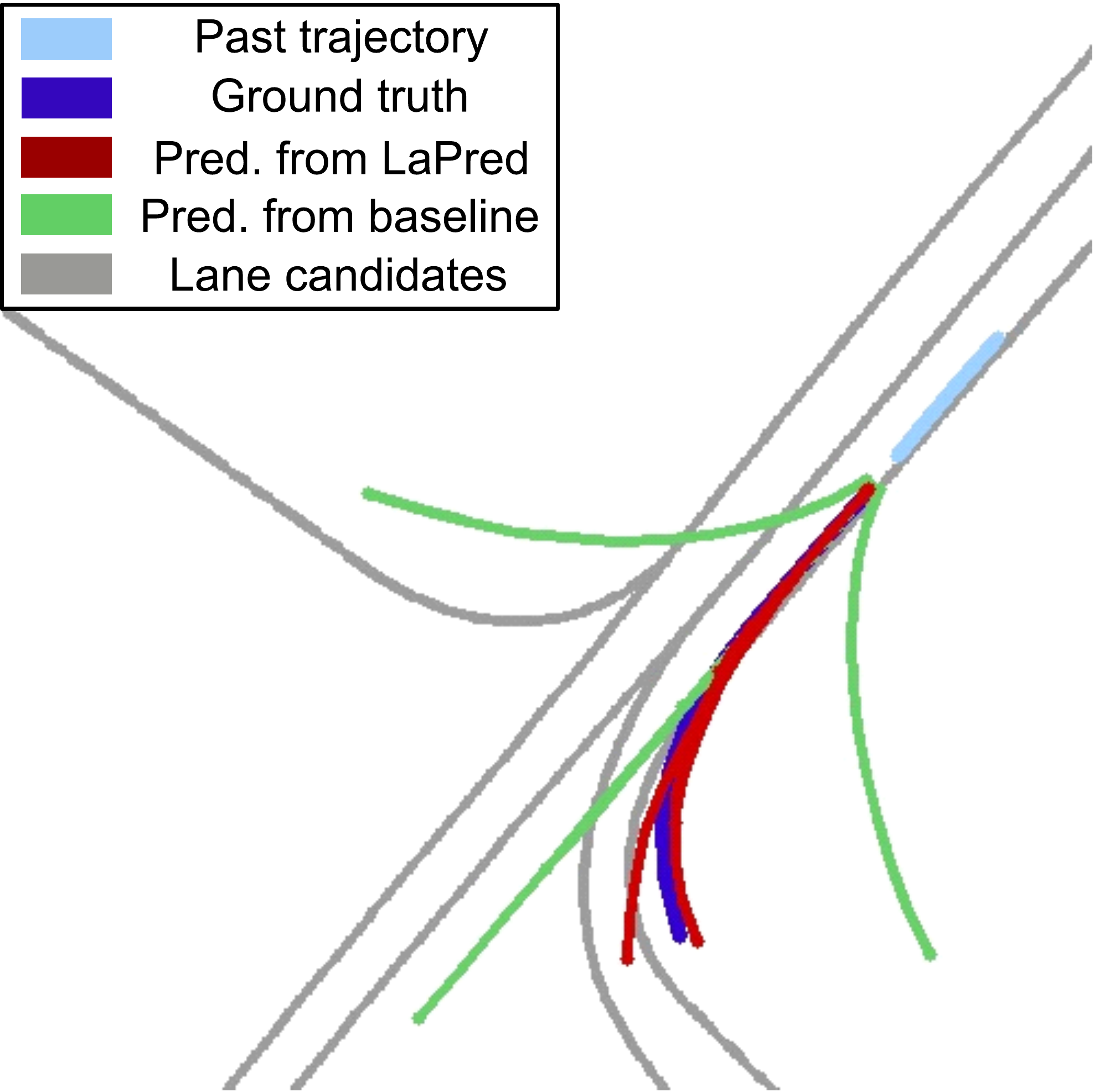}}
    
    \caption{Examples of the predicted trajectories obtained with LaPred and baseline on nuScenes dataset: (a)-(d) $K=1$ and (e)-(h) $K=5$. }
    \label{fig:illustration}
\vspace{-0.3cm}
\end{figure*}

Table \ref{table:ablation} presents the contributions of each idea to the performance gain achieved by LaPred over the baseline.  Method A is the baseline, which predicts the future trajectory based on only the past trajectory of the target agent using the CNN-LSTM.
    %\item Method A : A method to use only past trajectory of the target agent without lane information (Baseline)
     Method B takes lane information as additional input and uses hard selection for feature aggregation.
    In Method C, the loss function $L_{\rm lane-off}$ is used to train the model.
     Method D uses the past trajectories of the nearby agents on top of Method C.
     Method E uses the soft selection for the weighted feature aggregation.

We evaluate four performance metrics for the methods. As we add the additional ideas to the baseline model, we see that the prediction accuracy improves. Note that using lane information used in Method B offers the largest performance improvement. When the trajectories of the nearby agents are used along with lane information, further performance improvement is observed, which shows that the proposed LaPred successfully extracts the representation of surrounding environments to enhance the prediction accuracy. We also note that soft feature combining used in Method E enables non-negligible performance gain over hard feature combining. This shows that consideration of diverse lane candidates is more helpful than using a single best lane candidate in predicting the future trajectory. We also observe that the loss function $L_{\rm lane-off}$ used in the method C  contributes to the performance gain achieved by the proposed LaPred.

\subsubsection{Quantitative Results}
We evaluate the performance of the proposed model on nuScenes and Argoverse datasets.
For each dataset, we compare the proposed method with the existing methods that have already published performance results in the literature. Table \ref{table:nuscenes} presents the results on the nuScenes validation set.
We tried $K=1, 5, 10,$ and $15$ for performance evaluation.
The proposed LaPred model outperforms other prediction methods on most of the metrics considered.
For the metrics $FDE_{10}$, $ADE_{15}$, and $FDE_{15}$,  the proposed method is slightly worse than MHA-JAM \cite{trivedi_mha}, but the performance improvement for the rest is significant.
This result shows that using the instance-level lane information is  advantageous in capturing the scene context and interaction with the nearby agents over using 2D raster images.

Table \ref{table:argoverse} presents the results on the Argoverse validation set with $K =1,5,6,$ and $12$. The proposed method achieves the performance gain over the competitors in $ADE_5$, $FDE_5$, $ADE_6$, $FDE_6$, and $ADE_{12}$ metrics.
In the metrics $ADE_1$, $FDE_1$, and $FDE_{12}$, the performance of LaPred is comparable to that of the top-performing methods.
In particular, the proposed method achieves remarkable performance gain for $K=5,6$.  
From this, we deduce that the impact of lane information to find the meaningful modes from the trajectory distribution is maximized in these configurations. MTPLA is the algorithm that also uses instance-level lane information. It performs hard selection of a reference lane, whereas the proposed method applies the attention-based soft selection to consider multiple lane candidates to predict the future trajectory. Note that the proposed method demonstrates its advantage over MTPLA in $ADE_6$ and $FDE_6$.

\subsubsection{Prediction Examples}
Fig. \ref{fig:illustration} illustrates the samples of prediction produced by LaPred compared with the baseline for particular scenarios selected from the nuScenes dataset. Note that the baseline employs the encoder, which extracts the features only from the past trajectory of the target agent without using lane information. 
Fig. \ref{fig:illustration}  (a), (b), (c), and (d) present the trajectory samples generated with $K=1$ and the remaining figures show those with $K=5$.

We see from Fig. \ref{fig:illustration}  (a), (b), (c), and (d) that LaPred presents the predicted trajectory following one of the provided lanes well. On the contrary, the predictions from the baseline do not comply with the lane structure provided by the map. From the results, we confirm that the lane information provides important cues for improving the prediction accuracy, especially in the lateral direction.

In Fig. (e), (f), (g) and (h), both methods produce the five trajectory hypotheses to reflect multi-modal trajectory distribution.
We observe that the LaPred successfully generates multiple trajectories that are associated with different lane candidates. We note that lane information provides the target agent with a choice of admissible routes and LaPred exploits it to produce the diverse trajectories meaningful in a traffic context. Without lane information, the baseline produces inadequate prediction results in traffic conditions. To overcome this issue, the baseline is required to generate more trajectory samples.

\section{Conclusions} \label{conclusions}
In this paper, we proposed a new multi-modal trajectory prediction method that uses instance-level lane information.
The proposed LaPred captures the relation between the lanes and the trajectories of the agents through the trajectory-lane features obtained for each lane candidate. These trajectory-lane features are fused with the appropriate weights obtained from the task of identifying the reference lane. Based on the joint representation of the surrounding environment found by the encoder, the decoder generates multi-modal future trajectories compliant with the lane structure.
We trained our model with a self-supervised learning loss, which guides our model to identify the reference lane among $N$ lane candidates.
The experiments conducted on nuScenes and Argoverse datasets demonstrated that the LaPred achieved state-of-the-art performance in some metrics and produced the reasonable prediction results in challenging traffic scenarios.

\section{Acknowledgement}
This work was supported in part by Autonomous Driving Center, R\&D Division, Hyundai Motor Company, the Institute of Information \& Communications Technology Planning \& Evaluation (IITP) grant funded by the Korea government (MSIT) (No. 2020-0-01373, Artificial Intelligence Graduate School Program (Hanyang University)) and the National Research Foundation of Korea grant funded by the Korea government (MSIT) (No. 2020R1A2C2012146).

{\small
\bibliographystyle{ieee_fullname}
\bibliography{egbib}
}

\newpage
\clearpage
\onecolumn
\appendix
\section*{Supplementary}

In this supplementary material, we present the implementation details of the proposed LaPred method.

\section{System Setup and Training Details}

\begin{itemize}
    \item Time steps of past trajectory
    \begin{itemize}
        \item nuScenes dataset: $\tau=4$ (2 seconds)
        \item Argoverse dataset: $\tau=20$ (2 seconds)
    \end{itemize}
    \item Time steps of future trajectory
    \begin{itemize}
        \item nuScenes dataset: $h=12$ (6 seconds)
        \item Argoverse dataset: $h=30$ (3 seconds)
    \end{itemize}
    \item Number of lane candidates: $N=6$
    \item Setting for lane candidates
    \begin{itemize}
        \item nuScenes dataset
        \begin{itemize}
            \item Length of lane candidate: 130m (Forward: 100m, Backward: 30m)
            \item Distance between adjacent coordinate points: 0.5m
            \item Total number of the coordinate points for each lane candidate: $M=260$
        \end{itemize}
        \item Argoverse dataset
        \begin{itemize}
            \item Length of lane candidate: 80m (Forward: 50m, Backward: 30m)
            \item Distance between adjacent coordinate points: 1.0m
            \item Total number of the coordinate points for each lane candidate:  $M=80$
        \end{itemize}
    \end{itemize}
    \item Adam optimizer with initial learning rate 0.0003
    \item Learning rate decay: Reduced by half when the validation loss is plateaued for more than three epochs
    \item Batch size: $B=32$
    \item Loss parameters: $\alpha=0.3$ and $\beta=0.7$
    
\end{itemize}
\clearpage

\section{Network Architecture}
Table \ref{fig:tfe} to \ref{fig:mtp} present the detailed network architectures of the TFE block, LA block, and MTP block, respectively.

\begin{table}[H]
\centering
\begin{tabular}{c|c|c|c|c|c|c}
\hline \hline 
Input                  & \multicolumn{2}{c|}{$V^{(P)}$} & \multicolumn{2}{c|}{$L^i$} & \multicolumn{2}{c}{$V^i$} \\ \hline
\multirow{12}{*}{1D CNN} & \multirow{2}{*}{Input size} & \multirow{2}{*}{$B \times \tau \times 2$} &
                        Input size & $B \times M \times 2$ & \multirow{2}{*}{Input size} &
                        \multirow{2}{*}{$B \times \tau \times 2$} \\
                        
                        &  &  & Specification & $u64-k3-s1-p1$ &  &  \\
                        
                        & \multirow{2}{*}{Specification} & \multirow{2}{*}{$u64-k2-s1-p0$} &
                        Output size & $B \times M \times 64$ &
                        \multirow{2}{*}{Specification} & \multirow{2}{*}{$u64-k2-s1-p0$} \\ \cline{4-5}
                        
                        &  &  & Input size & $B \times M \times 64$ &  & \\
                        
                        & \multirow{2}{*}{Output size} & \multirow{2}{*}{$B \times \left( \tau-1 \right) \times 64$} &
                        Specification & $u64-k3-s1-p1$ &
                        \multirow{2}{*}{Output size} & \multirow{2}{*}{$B \times \left( \tau-1 \right) \times 64$} \\
                        
                        &  &  &  Output size & $B \times M \times 64$ &  & \\ \cline{2-7}
                        
                        & \multirow{2}{*}{Input size} & \multirow{2}{*}{$B \times \left( \tau-1 \right) \times 64$} &
                        Input size & $B \times M \times 64$ &
                        \multirow{2}{*}{Input size} & \multirow{2}{*}{$B \times \left( \tau-1 \right) \times 64$} \\
                        
                        &  &  & Specification & $u96-k3-s1-p1$ &  & \\
                        
                        & \multirow{2}{*}{Specification} & \multirow{2}{*}{$u64-k2-s1-p0$} &
                        Output size & $B \times M \times 96$ &
                        \multirow{2}{*}{Specification} & \multirow{2}{*}{$u64-k2-s1-p0$} \\ \cline{4-5}
                        
                        &  &  & Input size & $B \times M \times 96$ &  & \\
                        
                        & \multirow{2}{*}{Output size} & \multirow{2}{*}{$B \times \left( \tau-2 \right) \times 64$} &
                        Specification & $u96-k3-s1-p1$ &
                        \multirow{2}{*}{Output size} & \multirow{2}{*}{$B \times \left( \tau-2 \right) \times 64$} \\
                        
                        &  &  & Output size & $B \times M \times 64$ &  & \\ \hline
                        
\multirow{3}{*}{LSTM}   & Input size & $B \times \left( \tau-2 \right) \times 64$ &
                        Input size & $B \times M \times 96$ &
                        Input size & $B \times \left( \tau-2 \right) \times 64$ \\
                        
                        & Specification & $u512$ &
                        Specification & $u2048$ &
                        Specification & $u512$ \\
                        
                        & Output size & $B \times 512$ &
                        Output size & $B \times 2048$ &
                        Output size & $B \times 512$ \\ \hline
                        
Output    & \multicolumn{2}{c|}{$\xi_{V_p}$} & \multicolumn{2}{c|}{$\xi_{L^i}$} & \multicolumn{2}{c}{$\xi_{V^i}$} \\ \hline
                        & \multicolumn{6}{c}{Concatenation} \\ \hline
\multirow{12}{*}{FC}    & \multicolumn{3}{c|}{Input size} & \multicolumn{3}{c}{$B \times 3072$} \\
                        & \multicolumn{3}{c|}{Specification} & \multicolumn{3}{c}{$u2048$} \\
                        & \multicolumn{3}{c|}{Output size} & \multicolumn{3}{c}{$B \times 2048$} \\ \cline{2-7}
                        & \multicolumn{3}{c|}{Input size} & \multicolumn{3}{c}{$B \times 2048$} \\
                        & \multicolumn{3}{c|}{Specification} & \multicolumn{3}{c}{$u2048$} \\
                        & \multicolumn{3}{c|}{Output size} & \multicolumn{3}{c}{$B \times 2048$} \\ \cline{2-7}
                        & \multicolumn{3}{c|}{Input size} & \multicolumn{3}{c}{$B \times 2048$} \\
                        & \multicolumn{3}{c|}{Specification} & \multicolumn{3}{c}{$u1024$} \\
                        & \multicolumn{3}{c|}{Output size} & \multicolumn{3}{c}{$B \times 1024$} \\ \cline{2-7}
                        & \multicolumn{3}{c|}{Input size} & \multicolumn{3}{c}{$B \times 1024$} \\
                        & \multicolumn{3}{c|}{Specification} & \multicolumn{3}{c}{$u1024$} \\
                        & \multicolumn{3}{c|}{Output size} & \multicolumn{3}{c}{$B \times 1024$} \\ \hline
                 Output       & \multicolumn{6}{c}{$\xi^i$} \\ \hline \hline

\end{tabular}
\caption{Detailed network architecture of TFE block. $ux-ky-sz-pw$ represents a layer with the number of unit $x$, kernel size $y$, stride $z$, and padding on all side $w$.}
\label{fig:tfe}
\end{table}

\clearpage

\begin{minipage}{\textwidth}
\begin{minipage}[h]{0.48\textwidth}
\makeatletter\def\@captype{table}
\centering
\begin{tabular}{c|c|c}
\hline \hline
Input                   & \multicolumn{2}{c}{$\xi^{1:N}$} \\ \hline
                        & \multicolumn{2}{c}{Concatenation} \\ \hline
\multirow{21}{*}{FC}    & Input size & $B \times \left( 1024 \times N \right)$ \\
                        & Specification & $u512$ \\
                        & Output size & $B \times 512$ \\ \cline{2-3}
                        & Input size & $B \times 512$ \\
                        & Specification & $u512$ \\
                        & Output size & $B \times 512$ \\ \cline{2-3}
                        & Input size & $B \times 512$ \\
                        & Specification & $u256$ \\
                        & Output size & $B \times 256$ \\ \cline{2-3}
                        & Input size & $B \times 256$ \\
                        & Specification & $u256$ \\
                        & Output size & $B \times 256$ \\ \cline{2-3}
                        & Input size & $B \times 256$ \\
                        & Specification & $u64$ \\
                        & Output size & $B \times 64$ \\ \cline{2-3}
                        & Input size & $B \times 64$ \\
                        & Specification & $u64$ \\
                        & Output size & $B \times 64$ \\ \cline{2-3}
                        & Input size & $B \times 64$ \\
                        & Specification & $u(N)$ \\
                        & Output size & $B \times N$ \\ \hline
                        & \multicolumn{2}{c}{Softmax} \\ %\cline{2-3}\\
                        %& \multicolumn{2}{c}{$p_{1:N}$}
                        %& \multicolumn{2}{c}{$w_{1:N}$}
\hline \hline
\end{tabular}
\caption{Detailed network architecture of LA block.}
\label{fig:la}

\end{minipage}
% \qquad
%\centering
\begin{minipage}[t]{0.48\textwidth}
\makeatletter\def\@captype{table}
\centering
\begin{tabular}{c|c|c}
\hline \hline
Input                   & $\xi$ & $\xi_{V_p}$ \\ \hline
                        & \multicolumn{2}{c}{Concatenation} \\ \hline
\multirow{9}{*}{FC$(k)$}     & Input size & $ B \times 1536 $ \\
                        & Specification & $u512$ \\
                        & Output size & $B \times 512$ \\ \cline{2-3}
                        & Input size & $B \times 512$ \\
                        & Specification & $u512$ \\
                        & Output size & $B \times 512$ \\ \cline{2-3}
                        & Input size & $B \times 512$ \\
                        & Specification & $u256$ \\
                        & Output size & $B \times 256$ \\ \hline
\multirow{6}{*}{Shared FC}     & Input size & $B \times 256$ \\
                        & Specification & $u256$ \\
                        & Output size & $B \times 256$ \\ \cline{2-3}
                        & Input size & $B \times 256$ \\
                        & Specification & $u\left( h \times 2 \right)$ \\
                        & Output size & $B \times \left( h \times 2 \right)$ \\ \hline
Output                  & \multicolumn{2}{c}{$\hat{V}^{(f,k)}$} \\
\hline \hline
\end{tabular}
\caption{Detailed network architecture of MTP block.}
\label{fig:mtp}

\end{minipage}
\end{minipage}

\section{Single-agent vs. Multi-agent features for employing nearby agents.}
Table \ref{table:agts} presents the comparison of the prediction performance among different methods to employ nearby agents in the TFE block. Specifically, we compare our proposed method $LaPred_{SL}$, which considers a single agent per lane, against two different methods $LaPred_{ML}$ and $LaPred_{M}$, which aggregate multiple nearby agents per lane by using max-pooling. In $LaPred_{ML}$, the closest lane to each agent is identified then the agent features that share the same closest lane are aggregated to their corresponding closest lane. On the other hand, $LaPred_{M}$ aggregates every agent feature to all lanes regardless of their distances to lanes.

\begin{table}[tbh]
\centering
\begin{tabular}{c|cccccccc}
Method & \multicolumn{1}{c}{$ADE_5$} & \multicolumn{1}{c}{$FDE_5$} & \multicolumn{1}{c}{$ADE_{10}$} & \multicolumn{1}{c}{$FDE_{10}$} \\ \hline
$LaPred_{SL}$ & 1.53 & 3.37 & 1.21 & 2.61 \\
$LaPred_{ML}$ & 1.56 & 3.42 & 1.22 & 2.63 \\
$LaPred_{M}$ & 1.60 & 3.56 & 1.25 & 2.68 \\
\end{tabular}
\vspace*{-4mm}
\caption{Performance of different methods to aggregate nearby agents evaluated on nuScenes validation set.}
\label{table:agts}
\end{table}

\end{document}